\def\year{2019}\relax
\documentclass[letterpaper]{article} 
\usepackage{custom}  
\usepackage{times}  
\usepackage{helvet}  
\usepackage{courier}  
\usepackage{url}  
\usepackage{graphicx}  
\usepackage{amsmath}
\usepackage{tikz}
\usepackage{amssymb}
\usepackage[linesnumbered,ruled]{algorithm2e}
\usepackage{algpseudocode}
\usepackage{subcaption}

\newcommand{\act}[1]{#1}

\newcommand{\pref}{\succ}
\newcommand{\state}[1]{#1}

\newcommand{\Cp}{\ensuremath C}
\newcommand{\ub}[1]{\ensuremath #1^<}

\newcommand{\Yolobot}{\textsc{Yolobot}}
\newcommand{\omcts}{\mbox{O-MCTS}}
\newcommand{\pbmcts}{\mbox{PB-MCTS}}
\newcommand{\nmcts}{\mbox{N-MCTS}}
\newcommand{\ormdp}{\mbox{OR-MDP}}
\newcommand{\mixmax}{\textsc{MixMax}}

\newcommand{\game}[1]{\textsl{#1}}

\usepackage{rotating}
\usepackage{multirow}
\usepackage{longtable}
\usepackage{colortbl}
\usepackage{booktabs}
\usepackage{url}
\usepackage{todonotes}
\newcommand{\citet}[1]
{\citeauthor{#1}˜\shortcite{#1}}
\newcommand{\citep}{\cite}

\newcommand{\citealt}[1]
{\citeauthor{#1}~\citeyear{#1}}

\begin{document}
%
\title{Ordinal Monte Carlo Tree Search}
\author{Tobias Joppen \and
Johannes F\"urnkranz\\
Technische Universit\"at Darmstadt, Germany\\
\{tjoppen,juffi\}@ke.tu-darmstadt.de}

\maketitle
\begin{abstract} 
In many problem settings, most notably in game playing, an agent receives a possibly delayed reward for its actions.
Often, those rewards are handcrafted and not naturally given.
Even simple terminal-only rewards, like winning equals $1$ and losing equals $-1$, can not be seen as an unbiased statement, since these values are chosen arbitrarily, and the behavior of the learner may change with different encodings, such as setting 
the value of a loss to $-0.5$, which is often done in practice to encourage learning.
It is hard to argue about good rewards and the performance of an agent often depends on the design of the reward signal. 
%
In particular, in  
domains where states by nature only have an ordinal ranking and where meaningful distance information between game state values is not available,
a numerical reward signal is necessarily biased.
In  this paper we take a look at Monte Carlo Tree Search (MCTS), a popular algorithm to solve MDPs, highlight a reoccurring problem concerning its use of rewards, and show that an ordinal treatment of the rewards overcomes this problem.
Using the General Video Game Playing framework we show dominance of our newly proposed ordinal MCTS algorithm over preference-based MCTS, vanilla MCTS and various other MCTS variants.

\end{abstract}

\section{Introduction}
\label{sec:Introduction}

In reinforcement learning, an agent solves a Markov decision process (MDP) by selecting actions that maximize its long-term reward.
Most state-of-the-art algorithms 
assume numerical rewards.
In domains like finance, real-valued reward is naturally given,
but many other domains do not have a natural numerical reward representation.
In such cases, numerical values are often handcrafted by experts so that they optimize the performance of their algorithms. 
This process is not trivial, and it is hard to argue about good rewards.
Hence, such handcrafted rewards may easily be erroneous and contain biases.
For special cases such as domains with true ordinal rewards, it has been shown that it is impossible to create numerical rewards that are not biased. For example,
\cite{yannakakis2017ordinal} argue that emotions need to be treated as ordinal information. 

In fact, it often is hard or impossible to tell whether domains are real-valued or ordinal by nature.
Experts may even design handcrafted numerical reward without thinking about alternatives, since using numerical reward is state of the art and most algorithms need them.
In this paper we want to emphasize that numerical rewards do not have to be the ground truth and it may be worth-while for the machine learning community to have a closer look on other options, ordinal being only one of them.

\paragraph{MCTS}
Monte Carlo tree search (MCTS) is a popular algorithm to solve MDPs.
MCTS is used in many successful AI systems, such as AlphaGo \cite{silver2017mastering} or top-ranked algorithms in the general video game playing competitions \cite{perez2018general,YOLOBOT}.
A reoccurring problem of MCTS is its behavior in case of danger:
As a running example we look at a generic platform game, where an agent has to jump over deadly gaps to eventually reach the goal at the right.
Dying is very bad, and the more the agent proceeds to the right, the better.
The problem occurs by comparing the actions \emph{jump} and \emph{stand still}
:
jumping either leads to a better state than before because the agent proceeded to the right by successfully jumping a gap, or to the worst possible state (\emph{death}) in case the jump attempt failed. Standing still, on the other hand, safely avoids death, but will never advance to a better game state.
MCTS averages the obtained rewards gained by experience, which lets it often choose the safer action and therefore not progress in the game, 
because the (few) experiences ending with its death pull down the average reward of \emph{jump} below the mediocre but steady reward of standing still.
Because of this, the behavior of MCTS has also been called \emph{cowardly} in the literature \cite{jacobsen2014monte,khalifa2016modifying}.

Transferring those platform game experiences into an ordinal scale eliminates the need of meaningful distances. 
In this paper, 
we present an algorithm that only depends on pairwise comparisons in an ordinal scale, and selects \emph{jump} over \emph{stand still} if it more often is better than worse.
We call this algorithm Ordinal MCTS (\omcts{}) and compare it to different MCTS variants using the General Video Game AI (GVGAI) framework \cite{perez2016general}.

%

In the next section we introduce MDPs, MCTS and some of its variants.
In Section~\ref{sec:O-MCTS}, we present our \omcts{} algorithm, followed by experiments (Sections~\ref{sec:Setup} and~\ref{sec:Experiments}) and a conclusion (Section~\ref{sec:Conclusion}).

\section{Monte Carlo Tree Search}
\label{sec:Foundations}

In this section, we briefly recapitulate Monte Carlo tree search and some of its variants, which are commonly used for solving Markov decision processes.

\subsection{Markov Decision Process}
\label{sec:MDPs}
A \emph{Markov decision process} (MDP; \citealt{MDP}) can be formalized as quintuple
($S$, $A$, $\delta$, $r$, $\mu$) 
where
$S$ is the set of possible states $s$, $A$ the set of actions $a$ the agent can perform (with the possibility of only having a subset of possible actions $A_s\subset A$ available in state $s$), $\delta(s'|s,a)$ a state transition function, $r(s)\in \mathbb{R}$ a reward function  for reaching state $s$, and $\mu(s)\in [0,1]$  a distribution for starting states.
\citet{weng2011markov} has extended this notion to \emph{ordinal reward MDPs} (\ormdp{}),
where rewards are defined over a qualitative, ordinal scale $O$, in which states can only be compared to a obtain a preference between them, but the feedback does not provide any numerical information which allows to assess a magnitude of the difference in their evaluations.

The goal is 
learn a \emph{policy} $\pi(\act{a}\mid \state{s})$ that defines the probability of selecting an action $\act{a}$ in state $\state{s}$. The optimal policy $\pi^*(\act{a}\mid \state{s})$ maximizes the expected, cumulative reward in the MDP setting \cite{ReinforcementLearning}, or the preferential information for each reward in a trajectory in the \ormdp{} setting \cite{weng2011markov}.
For finding an optimal policy, one needs to solve the so-called exploration/exploitation problem. The state/action spaces are usually too large to sample exhaustively. Hence, it is required to trade off the improvement of the current, best policy (exploitation) with an exploration of unknown parts of the state/action space.

\subsection{Monte Carlo Tree Search}
\label{sec:MCTS}
Monte Carlo tree search (MCTS) is a method for approximating an optimal policy for a MDP.
It builds a partial search tree, which is more detailed where the rewards are high.
MCTS spends less time evaluating less promising action sequences, but does not avoid them entirely in order to explore the state space.
The algorithm iterates over four steps \cite{MCTSSurvey}:

\begin{enumerate}
\item \emph{Selection:} Starting from the root node $v_0$ which corresponds to start state $\state{s}_0$, a \emph{tree policy} traverses to deeper nodes $v_k$, until a state with unvisited successor states is reached.
\item \emph{Expansion:} One successor state is added to the tree.
\item \emph{Simulation:} Starting from the new state, a so-called \emph{rollout} is performed, i.e., random actions are played until a terminal state is reached or a depth limit is exceeded.
\item \emph{Backpropagation:} The reward of the last state of the simulation is backed up through the selected nodes in tree.
\end{enumerate}
The UCT formula
\begin{equation}
a^*_v = \max_{a \in A_v} \bar{X}_v(a) + 2 \Cp \sqrt{\frac{2 \ln n_v}{n_{v}(a)}}
\label{eq:uct}
\end{equation}
\noindent
is used to select the most interesting action $a^*_v$ in a node $v$ by trading off the expected reward estimated as $\bar{X}_v(a) =$ $\sum_{i=0}^{n_v}X_v^{(i)}(a)/n_v(a)$ from $n_v(a)$ samples $X_v^{(i)}(a)$ in which action $a$ has been taken in node $v$,
with 
an exploration term $2\cdot\sqrt{2 \ln (n_v)/n_v(a)}$. The trade-off parameter $\Cp$ is often set to $\Cp = 1/\sqrt{2}$, which has been shown to ensure convergence for rewards $\in [0,1]$ \cite{UCT}.

In the following, we will often omit the subscript $v$ when it is clear from the context.

\subsection{\mixmax{} Modification}
\label{sec:MixMax}
As mentioned in the introduction, MCTS has been blamed for cowardly behavior in the sense that it often prefers a safer, certain option over a more promising but uncertain outcome.
To change this behavior, \citet{jacobsen2014monte} proposed to use \mixmax, which uses a mix between the maximum and the average reward
\begin{equation}
 \bar{X}(a) = Q\cdot \max_{i}
 X^{(i)}(a) + (1-Q)\sum_{i=1}^{n(a)}\frac{X^{(i)}(a)}{n(a)},
\label{eq:mixmax}
\end{equation}
where $Q$ is a parameter to trade off between the two values.
As illustrated further below (Figure~\ref{fig:loss}), this is a possible way to encourage MCTS to boost actions that can lead to high rated states.
Hence, \mixmax{} may solve the running problem given a well-tuned $Q$ value.

The benefit of \mixmax{} is its simplicity which makes it very cheap to compute. However, the use of the maximum makes does not take into account the distribution of rewards, which makes it very sensitive to noise: a single outlier may lead to a high \mixmax{} bonus.
Hence, MCTS may choose a generally very deadly action just because it survived once and got a good score for that, thereby, in a way, inverting the problem with the conservative action selection of MCTS. Also note that 
in comparison to vanilla MCTS, this bonus does not decrease with a higher number of deadly samples.

The \mixmax{} modification has already been used in the General Video Game Framework to reduce cowardly behavior. 
\citet{khalifa2016modifying}b
found that $Q=0.25$ exhibits more human-like behavior so that we will also use this 
parameter setting in our experiments.

\subsection{Preference-Based Monte Carlo Tree Search}
\label{sec:PB-MCTS}

A version of MCTS that uses preference-based feedback (\pbmcts{}) was recently introduced by \citet{joppen2018preference}. 
In 
this setting, the agent receives rewards in the form of preferences over states.
Hence, feedback about single states $s$ is not available, it can only be compared
to another state $s'$, i.e., $s\succ s'$ ($s$ dominates $s'$), $s'\succ s$, or $s\not\sim s'$ ($s$ and $s'$ are incomparable).

An iteration of \pbmcts{} contains the same abstract steps like MCTS, but their realization differs.
First and foremost, it is impossible to use preference information on a vanilla MCTS iteration, since it only samples a single trajectory, whereas a second state
 is needed for a comparison.
Hence, \pbmcts{} does not select a single path per iteration but an entire subtree of the search tree. 
In each of its nodes, 
two actions are selected 
that can be compared to each other.
In the backpropagation phase, the two selected actions in a node both have at least one trajectory.
All trajectories are compared and the received preference information is stored.
For the selection step, 
a modified version of the dueling bandit algorithm RUCB \cite{zoghi14} is used to select two actions per node given the stored preferences.

%
%

\begin{figure}[t]
\centering
\includegraphics[width=\columnwidth]{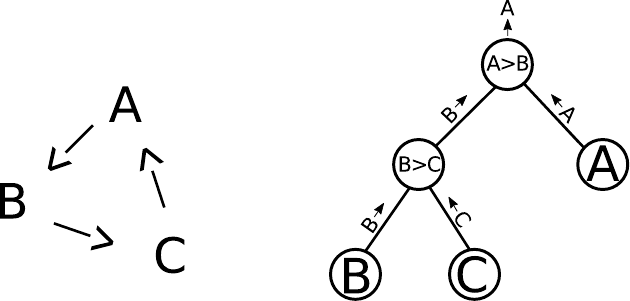}
\caption{Three nontransitive actions. The tree introduces a bias to solve nontransitivity.}
\label{fig:transitivity}
\end{figure}

There are two 
main disadvantages with this approach:
\begin{enumerate}
	\item \emph{No transitivity} is used. Given ten actions $a_0$ to $a_9$, MCTS needs only at most $10$ iterations to have a first fair estimation of the quality of each of those $10$ actions. In the preference-based approach, each action has to be compared with each other action until a first complete estimation can be done. These are $(10 \cdot 9)/2 = 45$ iterations, i.e., in general the effort is quadratic in the number of actions. 
	\item A \emph{binary subtree} is needed to learn on each node of the currently best trajectory. Instead of a path of length $n$ for vailla MCTS, the subtree consists of $2^n-1$ nodes and $2^{n-1}$ trajectories instead of only one, 
	causing an 
	exponential blowup of \pbmcts{}'s search tree.
\end{enumerate}
Hence, we believe that \pbmcts{} does not make optimal use of available computing resources, since on a local perspective, transitivity information is lost, and on a global perspective, the desired asymmetric growth of the search tree is undermined by the need for selecting a binary tree.
Note that even in the case of a non-transitive domain, \pbmcts{} will nevertheless 
obtain a transitive policy, as illustrated in Figure~\ref{fig:transitivity}, where the circular preferences between actions \textsf{A}, \textsf{B}, and \textsf{C} can not be reflected in the resulting tree structure. 

\section{Ordinal Monte Carlo Tree Search}
\label{sec:O-MCTS}
In this section, we introduce \omcts, an MCTS variant which only relies on 
ordinal information to learn a policy.
We will first present the algorithm, and then take a closer look at the differences to MCTS and \pbmcts.

\subsection{O-MCTS}
\label{sec:O-MCTSAlg}
\emph{Ordinal Monte Carlo tree search} (\omcts) proceeds like conventional MCTS as introduced in Section~\ref{sec:MCTS}, but 
replaces the average value $\bar{X}_v(a)$ in~\eqref{eq:uct}
 with the Borda score $B_v(a)$ of an action $a$.
To calculate the Borda score for each action in a node, \omcts{} stores the backpropagated ordinal values, and estimates pairwise preference probabilities $P_v(a\pref b)$ from these data.
Hence, it is not necessary to do multiple rollouts in the same iteration as in \pbmcts{} because
current rollouts can be directly compared to previously observed ones.

Note that $P_v(a \succ b)$ can only be estimated if each action was visited at least once.
Hence, similar to other MCTS variants, we enforce this by always selecting non-visited actions in a node first.

\subsection{The Borda Score}
\label{sec:Borda}


The Borda score is based on the Borda count which has its origins in voting theory \cite{black1976partial}. Essentially, it estimates the probability of winning against a random competitor. In our case, $B_v(a)$ estimates 
the probability of action $a$ to win against any other action $a' \neq a$ available in node $v$.

To calculate the Borda score $B_v(a)$, we store all backpropagated ordinal values $o \in O$ for each action $a \in A_v$ available in node $v$.
A simple solution to summarize this information is to use a two-dimensional array $c_v: O \times A \rightarrow \mathbb{N}$ to count how often value $o$ is obtained by playing action $a$ in node $v$.
Given $c(o,a)$ in a node, we can derive the estimated density probabilities 
\begin{equation}
P(o\mid a)=\frac{c(o,a)}{\sum_{o'\in O}c(o',a)}.
\label{eq:Poa}
\end{equation}
for
receiving ordinal reward $o$ by playing action $a$ in this node.
%
The probability of receiving an ordinal reward worse than $o$ for action $a$ (which we denote with $\ub{o}$) is then
\begin{equation}
P(\ub{o} \mid a) = \sum_{o' \prec o \in O}P(o'\mid a).
\label{eq:ub}
\end{equation}
Given this, the probability $P(a \succ b)$ of action $a$ beating action $b$ can be estimated as
%
\begin{equation}
P(a\succ b) =  \sum_{o\in O} P(o\mid a) \left( 
P(\ub{o} \mid b)
+ \frac{1}{2} P(o \mid b) \right)
\label{eq:Pab}
\end{equation}
For each ordinal value $o$, this estimates the probability 
of $a$ receiving reward $o$ while $b$ receiving a lesser reward (plus half of the probability that $b$ receives the same reward to deal with ties).
This is then summed up over all possible values $o$.

The Borda score of $a$ is then the average win probability of $a$ over all other actions available in this node:
\begin{equation}
B(a) = \frac{1}{|A-1|}\sum_{b \neq a \in A} 
P(a \succ b).
\end{equation}
It has several properties that encourage its use as a value estimator:
%
\begin{enumerate}
	\item $B(a) = 1$ if and only if action $a$ strictly dominates any other action.
	 Action $a$ seems to be the best option and has to get the highest estimate:
\begin{equation*}
\begin{split}
B(a) = 1 \Leftrightarrow & \; \forall b \in A \setminus a : P(a \succ b) = 1 \\
\Leftrightarrow & \; \exists o\in O: 
\forall \hat{o} \prec 
o: P(\hat{o} \mid a)=0 \\
   & \qquad \qquad \wedge \forall \hat{o}\succeq o: P(\hat{o} \mid b)=0.
\end{split}
\end{equation*}
	 \item $B(a) = 0$ if and only if action $a$ is strictly dominated by any other action.
	 If an action is worse than any other action, it has to get the lowest possible estimate:
\begin{equation*}
\begin{split}
B(a) = 0 \Leftrightarrow & \; \forall b \in A \setminus a : P(a \succ b) = 0 \\
\Leftrightarrow & \; \exists o\in O : \forall \hat{o}\prec o: P(\hat{o} \mid b)=0 \\
& \qquad \qquad \wedge \forall \hat{o}\succeq o: P(\hat{o} \mid a)=0.
\end{split}
\end{equation*}
	 \item  $B(a) = B(a')$ if two actions $a$ and $a'$ have equal ordinal outcomes:
	 
Since $P(o \mid a)=P(o \mid a')$, and the remaining terms to compute $B(a)$ and $B(a')$ are all identical, $B(a) = B(a')$ must hold.
%
\end{enumerate}

\subsection{Incremental Update}
%
In order to reduce computation, we do not compute the counts $c_v(a,o)$ used in~\eqref{eq:Poa}, but maintain counts $c_v(a,\ub{o})$, from which $P_v(\ub{o}\mid a)$  can be directly estimated, thereby avoiding the summation in~\eqref{eq:ub}.
For each rollout which took action $a$ in node $v$ yielding a reward $o_a$, we increase the
counts $c_v(\ub{o},a)$ for all $o \prec o_a$, as well as the counters $n_v(a)$ and $n_v$. From this, 
 the Borda count can be updated incrementally in the backpropagation step:
Given a new ordinal reward $o_a$ for action $a$ in node $v$, the Borda count for all $|A(v)|$ actions of $v$ have to be updated.
For each action $b\neq a$ of $v$ we can update $P_v(a\succ b)$ as follows:
\begin{equation}
\begin{split}
P^{(t+1)}(a \succ b) \leftarrow & \; \alpha P^{(t)}(a\succ b) + \\
&(1-\alpha)(P^{(t+1)}(\ub{o_a} \mid b) + \\
&\frac{1}{2}P^{(t+1)}(o_a \mid b) ),
\end{split}
\end{equation}
where $\alpha = 
n_v^{(t)}(a)/n_v^{(t+1)}(a)$
is the relative proportion of data from time step $t$ such that all iterations are weighted equally.
%

\subsection{Differences to MCTS}
\label{sec:differences}
Although the changes from MCTS to \omcts{} are comparably small, the algorithms 
have very different characteristics.
In this section, we highlight some of the differences between \omcts{} and MCTS.

\paragraph{Loss function}
MCTS and \omcts{} do not use the same loss function.
Consider  UCT values 
at an example node $v$ with two actions $a$ and $b$.
The past rollouts were $a\leftarrow 0.1,1,0.1$ and $b\leftarrow 0.3,0.35,0.25$.

MCTS averages different backpropagated values and compares them directly.
This can be seen as minimizing the \emph{linear loss}.
Here $a$ is better: $\varnothing_a=0.4 > \varnothing_b=0.3$.
\omcts{} has a different loss function:
instead of averaging the values, a preference comparison is used, and 
the action is chosen, which dominates the other more frequently.
This can be seen as minimizing 
a \emph{ranking loss}.
Here $b$ is better, as $P(a \succ b)=\frac{1}{3}<\frac{2}{3}=P(b \succ a)$ because $b$ has more wins than $a$.
Which loss function should be used depends on the specific problem.

\paragraph{Ordinal Values Only}
The most prominent difference between MCTS and \omcts{} is that for problems where only ordinal reward exist, MCTS is not applicable without creating an artificial reward signal. Any assignment of numerical values to ordinal values is arbitrary and will add a bias
\cite{yannakakis2017ordinal}. Similarly, a linear loss function (or any other loss function that uses value differences) will also introduce a bias, and a ranking loss should be used instead.

\begin{figure}[t]
\centering
\includegraphics[width=\columnwidth]{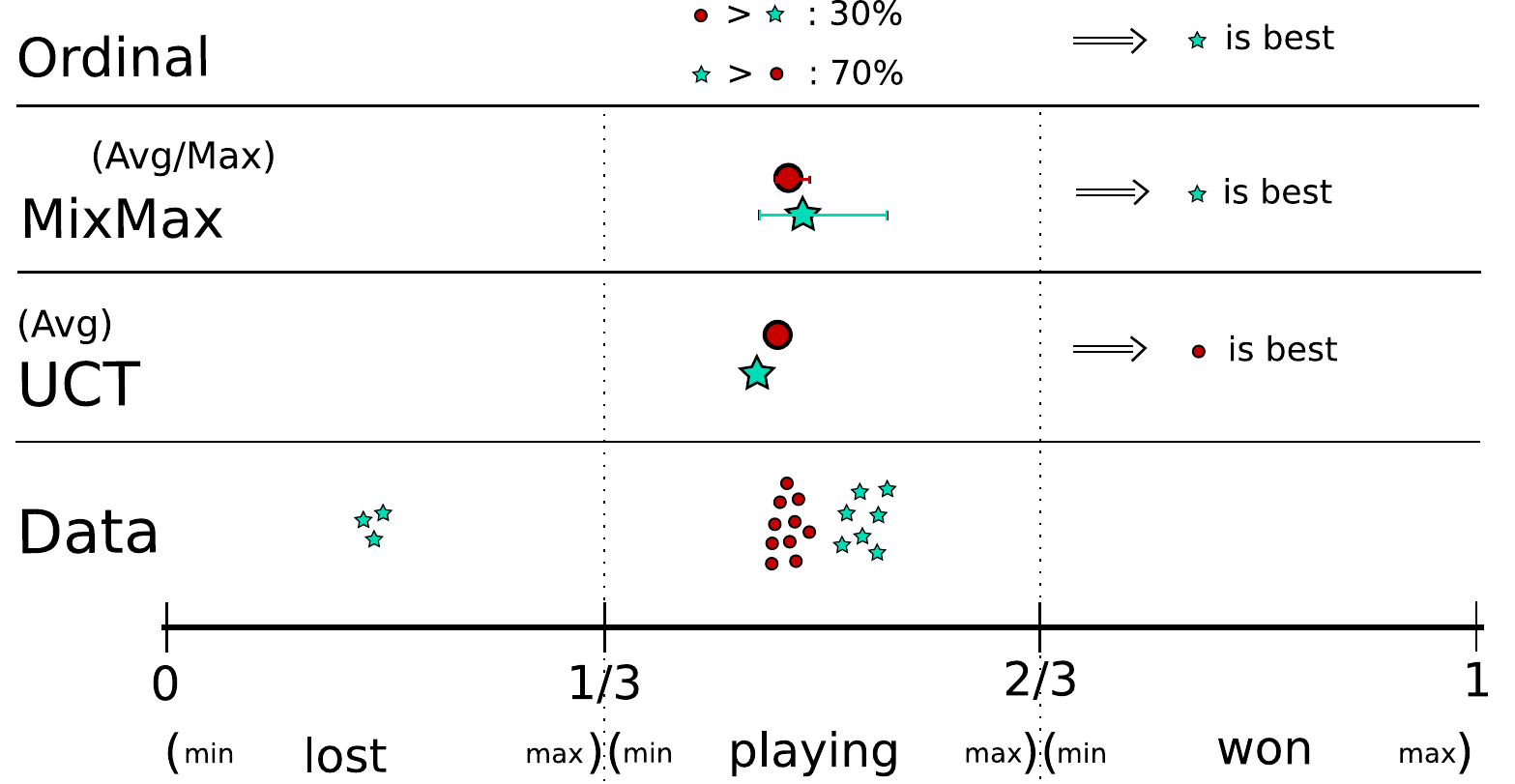}
\caption{Two actions with different distributions.}
\label{fig:loss}
\end{figure}

\paragraph{Cowardly Behavior}
As mentioned previously, MCTS has been blamed for behaving cowardly, by prefering
safe but unyielding actions over actions that have some risk but will in the long
run result in higher rewards.
As an example, consider Figure~\ref{fig:loss}, which shows in its bottom row the distribution of trajectory values for two actions over a range of possible rewards.
One action (circles) has a mediocre quality with low deviation, whereas
the other (stars) is sometimes worse but often better than the first one.
Since MCTS prioritizes the stars only if the average is above the average of circles, MCTS would often choose the safe, mediocre action.
In the literature one can find many ideas to tackle this problem, like \mixmax{} backups (cf.~Section~\ref{sec:MixMax}) or adding domain knowledge (e.g., by giving a direct bonus to actions that should be executed \cite{perez2018general,YOLOBOT}).
\omcts{} takes a different point of view, by not comparing average values but by comparing how often stars are the better option than circles and vice versa. As a result, it would prefer the circle action, which is preferable in 70\% of the games.

\paragraph{Normalization}
Although MCTS does not depend on normalized reward values, in practice they are nevertheless often normalize to a $[0,1]$ range in order to 
simplify the tuning of the $\Cp$ parameter.
%
%
\omcts{} is already normalized in the sense that all values $B_v(a)$ are in the range $[0,1]$. Note, however, that this a local, relative scaling and not a global, absolute scale as in regular MCTS. If $B_v(a) > B_w(b)$ this does not mean that $a$ is a better action than $b$ unless $v = w$.

MCTS can be modified to use local normalization as well by storing the minimal ($r_v^{min}$) and maximal ($r_v^{max}$) reward seen in each node $v$. 
For each new sample in $v$, these values are updated using the received reward $r$, which is then normalized using  $\bar{r}=(r-r_v^{min})/(r_v^{max}-r_v^{min})$.

In our experiments, we tested this version under the name of
\textsc{Normalized-MCTS} (\nmcts{}).

\paragraph{Computational Time}
Even though we propose an incremental update for the Borda score, it should be mentioned that calculating a running average (MCTS) is faster than calculating the Borda score (\omcts{}).
In our experiments, the Borda score needed $3$ to $20$ times more time than averaging
depending 
on the size of $O$ and $A$.

\section{Experimental Setup}
\label{sec:Setup}
We test the five algorithms described above (MCTS, \omcts{}, \nmcts{}, \mixmax{} and \pbmcts{}) using the General Video Game AI (GVGAI) framework \cite{perez2016general}.
GVGAI has implemented a variety of different video games and provides playing agents with a unified interface to simulate moves using a forward model.
Using this forward model is expensive so that simulations take a lot of time.
We use the number of calls to this forward model as a computational budget.
In comparison to using the real computation time, it is independent of specific hardware, algorithm implementations, and side effects such as logging data.

Our algorithms are given access to the following pieces of information provided by the framework:
\begin{enumerate}
	\item \emph{Available actions}: The actions the agent can perform in a given state
	\item \emph{Game score}: The score of the given state $\in \mathbb{N}$. Depending on the game this ranges from $0$ to $1$ or $-1000$ to $10000$.
	\item \emph{Game result}: The result of the game: \emph{won}, \emph{lost} or \emph{running}.
	\item \emph{Simulate action}: The forward model. It is stochastic, e.g., for enemy moves or random object spawns.
\end{enumerate}
\subsection{Heuristic Monte Carlo Tree Search}
The games in GVGAI have a large search space with $5$ actions and up to $2000$ turns.
Using vanilla MCTS, one rollout may use a substantial amount of time, since up to $2000$ moves have to be made to reach a terminal state.
To achieve a good estimate, many rollouts have to be simulated.
Hence it is common to stop rollouts early at non-terminal states, using a heuristic to estimate the value of these states.
In our experiments, we use this variation of MCTS, adding the maximal length for rollouts \textit{RL} as an additional parameter. 
The heuristic value at non-terminal nodes is computed in the same way as the terminal reward (i.e., it essentially corresponds to the score at this state of the game).
\subsection{Mapping Rewards to $\mathbb{R}$}
The objective function has two dimensions: on the one hand, the agent needs to win the game by achieving a certain goal, on the other hand, the agent also needs to maximize its score. 
Winning is more important than getting higher scores.

Since MCTS needs its rewards being $\in \mathbb{R}$ or even better $\in [0,1]$, the two-dimensional target function needs to be mapped to one dimension, in our case for comparison and ease of tuning parameters into $[0, 1]$.
%
%
Knowing the possible scores of a game, the score can be normalized by $r_{norm} = (r-r_{min})/(r_{max}-r_{min})$ with $r_{max}$ and $r_{min}$ being the highest and lowest possible score. Note that this differs from the \nmcts{} normalization discussed in Section~\ref{sec:differences} in that here global extrema are used, whereas \nmcts{} uses the extrema seen in each node.

For modeling the relation  \emph{lost} $\prec$ \emph{playing} $\prec$ \emph{won} which must hold for all states, we split 
the interval $[0,1]$ into three equal parts 
(cf.~also the axis of Figure~\ref{fig:loss}):

  \begin{equation}
  	r_{mcts} = \frac{r_{norm}}{3} + 
    \begin{cases}
      0, & \text{if \it lost} \\
      \frac{1}{3}, & \text{if \it playing} \\
      \frac{2}{3}, & \text{if \it won}.
    \end{cases}
    \label{eq:rewards}
  \end{equation}
This is only one of many possibilities to map the rewards to $[0,1]$, but it is an obvious and straight-forward approach. 
Naturally, the results for the MCTS techniques, which use this reward, will change when a different reward mapping is used, and their results can probably be improved by shaping the reward. In fact, one of the main points of our work is to show that for 
%
\omcts{} (as well as for \pbmcts{}) no such reward shaping is necessary because these algorithms do not rely on the numerical information. In fact, for them, 
the mapped linear function with $a\succ b \Leftrightarrow r_{mcts}(a) > r_{mcts}(b)$ is equivalent to the preferences induced by the two-dimensional feedback.

\subsection{Selected Games}
GVGAI provides users with many games.
Doing an evaluation on all of them is not feasible.
Furthermore, some results would exhibit erratic behavior,
since the tested algorithms are not suitable for solving some of the games. For example, often true rewards are very sparse, and the agent has to be guided in some way to reliably 
solve the game.

For this reason, we manually played all the games and selected a variety of interesting, and not too complex games with different characteristics, which we believed to be solvable for the tested algorithms:
\begin{itemize}
	\item \game{Zelda}: The agent can hunt monsters and slay them with its sword. It wins by finding the key and taking the door.
	\item \game{Chase}: The agent has to catch all animals which flee from the agent. Once an animal finds a catched one, it gets angry and chases the agent. The agent wins once no more animal flee and loses if a chasing animal catches it.
	\item \game{Whackamole}: The agent can collect mushrooms which spawn randomly. A cat helps it in doing so. The game is won after a fixed amount of time or lost if the agent and cat collide.
	\item \game{Boulderchase}: The agent can dig through sand to a door that opens after it has collected ten diamonds. Monsters chase it through the sand turning sand into diamonds.
	\item \game{Surround}: The agent can win the game at any time by taking a specific action, or collect points by moving while leaving a snake-like trail. A moving enemy also leaves a trail. The game is lost if the agent collides with a trail.
	\item \game{Jaws}: The agent controls a submarine, which is hunted by a shark. It can shoot fish giving points and leaving an item behind. Once 20 items are collected, a collision with the shark gives a large number of points, 
	otherwise it
	 loses the game. Colliding with fish always loses the game. The fish spawn randomly on 6 specific positions.
	\item \game{Aliens}: The agent can only move from left to right and shoot upwards. Aliens come flying from top to bottom throwing rocks on the agent. For increasing the score, the agent can shoot the aliens or shoot disappearing blocks.
\end{itemize}

The number of iterations that can be performed by the algorithms depends
on the computational budget of calls to the forward model.
We tested the algorithms with $250$, $500$, $1000$ and $10000$ forward model uses (later called \emph{time resources}). Thus, in total, we experimented with $28$ problem settings ($7$ domains $\times$ $4$ time resources).

\subsection{Tuning Algorithms and Experiments}
All MCTS algorithms have two parameters in common, the
\emph{exploration trade-off} $\Cp$ and \emph{rollout length} $RL$.
For $RL$ we tested 4 different values: $5, 10, 25$ and $50$, and
for $\Cp$ we tested 9 values from $0$ to $2$ in steps of size $0.25$.
In total, these are 36 configurations per algorithm.
To reduce variance, we have repeated each experiment 40 times.
Overall,  
5 algorithms with 36 configurations were run 40 times on 28 problems, 
resulting in 201600 games played for tuning.

Additionally, we compare the algorithms to \textsc{Yolobot}, a highly competitive GVGAI agent that won several challenges \cite{YOLOBOT,perez2018general}.
\textsc{Yolobot} is able to solve games none of the other five algorithms can solve.
Note that \textsc{Yolobot} is designed and tuned to act within a 20ms time limit.
Scaling the time resources might not lead to better behavior.
Still it is added for sake of comparison and interpretability of strength.
For \textsc{Yolobot} each of the $28$ problems is played $40$ times, which leads to $1120$ additional games or $202720$ games in total.\footnote{For
anonymization, we added the agents as supplementary material. In case of acceptance, they will be made publicly available.}

We are mainly interested on 
how well the different algorithms perform on the problems, given optimal tuning per problem.
To give an answer, we show the performance of the algorithms per problem in percentage of wins and obtained average score.
We do a Friedmann test on average ranks of those data with a posthoc Wilcoxon signed rank test to test for significance \cite{demvsar2006statistical}.
Additionally, we show and discuss the performance of all parameter configurations.

\begin{table}[htbp]
\caption{The results of algorithms tuned per row.
}
\label{tbl:reslt}
\centering
\small
\tabcolsep=0.11cm
	\begin{tabular}{cc||r|r|r|r|r|r}
		\begin{sideways}Game\end{sideways}&\begin{sideways}Time\end{sideways}
		&\multicolumn{1}{c|}{\begin{sideways}\smaller \omcts{}\end{sideways}}
		&\multicolumn{1}{c|}{\begin{sideways}\smaller MCTS\end{sideways}}
		&\multicolumn{1}{c|}{\begin{sideways}\smaller \nmcts{}\end{sideways}}
		&\multicolumn{1}{c|}{\begin{sideways}\smaller \textsc{Yolo}-\end{sideways} \begin{sideways}\smaller \textsc{bot}\end{sideways}}
		&\multicolumn{1}{c|}{\begin{sideways}\smaller \pbmcts{}\end{sideways}}
		&\multicolumn{1}{c}{\begin{sideways}\smaller \mixmax\end{sideways}}
		\\ 
		\hline
		\hline
		\multirow{8}{*}{\begin{sideways}\game{Jaws}\end{sideways}}
		&\multirow{ 2}{*}{$10^4$}
		&\textbf{     100\%}&\textbf{     100\%}&\textbf{     100\%}&      27.5\%&      80.0\%&      67.5\% \\ 
		&&\textbf{    1083.8}&     832.7&     785.7&     274.7&     895.7&     866.8\\\cline{2-8}
		&\multirow{ 2}{*}{$10^3$}
		&      92.5\%&\textbf{     95.0\%}&      92.5\%&      35.0\%&      52.5\%&      65.0\% \\ 
		&&\textbf{    1028.2}&     958.9&     963.2&     391.0&     788.5&     736.4\\\cline{2-8}
		&\multirow{ 2}{*}{500}
		&      85.0\%&      90.0\%&\textbf{      97.5\%}&      65.0\%&      50.0\%&      52.5\% \\ 
		&&     923.4&    1023.1&\textbf{    1078.2}&     705.7&     577.6&     629.0\\\cline{2-8}
		&\multirow{ 2}{*}{250}
		&      85.0\%&      85.0\%&\textbf{      87.5\%}&      32.5\%&      37.5\%&      37.5\% \\ 
		&&\textbf{    1000.9}&     997.6&     971.9&     359.6&     548.8&     469.0\\
		\hline
		\hline
		\multirow{8}{*}{\begin{sideways}\game{Surround}\end{sideways}}
		&\multirow{ 2}{*}{$10^4$}
		&\textbf{     100\%}&\textbf{     100\%}&\textbf{     100\%}&\textbf{     100\%}&\textbf{     100\%}&\textbf{     100\%} \\ 
		&&\textbf{      81.5}&      71.0&      63.5&      81.2&      64.3&      57.6\\\cline{2-8}
		&\multirow{ 2}{*}{$10^3$}
		&\textbf{     100\%}&\textbf{     100\%}&\textbf{     100\%}&\textbf{     100\%}&\textbf{     100\%}&\textbf{     100\%} \\ 
		&&\textbf{      83.0}&      80.8&      75.2&      77.3&      40.8&      25.0\\\cline{2-8}
		&\multirow{ 2}{*}{500}
		&\textbf{     100\%}&\textbf{     100\%}&\textbf{     100\%}&\textbf{     100\%}&\textbf{     100\%}&\textbf{     100\%} \\ 
		&&\textbf{      84.6}&      61.8&      79.3&      83.3&      26.3&      17.3\\\cline{2-8}
		&\multirow{ 2}{*}{250}
		&\textbf{     100\%}&\textbf{     100\%}&\textbf{     100\%}&\textbf{     100\%}&\textbf{     100\%}&\textbf{     100\%} \\ 
		&&\textbf{      83.4}&      64.7&      55.2&      76.1&      14.3&      10.3\\
		\hline
		\hline
		\multirow{8}{*}{\begin{sideways}\game{Aliens}\end{sideways}}
		&\multirow{ 2}{*}{$10^4$}
		&\textbf{     100\%}&\textbf{     100\%}&\textbf{     100\%}&\textbf{     100\%}&\textbf{     100\%}&\textbf{     100\%} \\ 
		&&\textbf{      82.4}&      81.6&      81.2&      81.5&      81.8&      77.0\\\cline{2-8}
		&\multirow{ 2}{*}{$10^3$}
		&\textbf{     100\%}&\textbf{     100\%}&\textbf{     100\%}&\textbf{     100\%}&\textbf{     100\%}&\textbf{     100\%} \\ 
		&&      79.7&      78.4&      77.7&\textbf{      82.2}&      76.9&      76.4\\\cline{2-8}
		&\multirow{ 2}{*}{500}
		&\textbf{     100\%}&\textbf{     100\%}&\textbf{     100\%}&\textbf{     100\%}&\textbf{     100\%}&\textbf{     100\%} \\ 
		&&      78.0&      77.3&      78.6&\textbf{      81.1}&      77.2&      76.0\\\cline{2-8}
		&\multirow{ 2}{*}{250}
		&\textbf{     100\%}&\textbf{     100\%}&\textbf{     100\%}&\textbf{     100\%}&\textbf{     100\%}&\textbf{     100\%} \\ 
		&&      77.7&      77.1&      77.1&\textbf{      79.3}&      75.8&      74.8\\
		\hline
		\hline
		\multirow{8}{*}{\begin{sideways}\game{Chase}\end{sideways}}
		&\multirow{ 2}{*}{$10^4$}
		&\textbf{      87.5\%}&      80.0\%&      80.0\%&      50.0\%&      67.5\%&      37.5\% \\ 
		&&\textbf{       6.2}&       6.0&       5.8&       4.8&       5.2&       3.9\\\cline{2-8}
		&\multirow{ 2}{*}{$10^3$}
		&      60.0\%&      50.0\%&      47.5\%&\textbf{      70.0\%}&      30.0\%&      17.5\% \\ 
		&&       4.8&       4.8&       5.0&\textbf{       5.1}&       3.7&       2.6\\\cline{2-8}
		&\multirow{ 2}{*}{500}
		&      55.0\%&      45.0\%&      45.0\%&\textbf{      90.0\%}&      27.5\%&      12.5\% \\ 
		&&       4.9&       4.5&       4.7&\textbf{       5.5}&       2.9&       2.1\\\cline{2-8}
		&\multirow{ 2}{*}{250}
		&      40.0\%&      32.5\%&      32.5\%&\textbf{      90.0\%}&      17.5\%&       7.5\% \\ 
		&&       4.2&       4.1&       4.2&\textbf{       5.6}&       2.5&       2.6\\
		\hline
		\hline
		\multirow{8}{*}{\begin{sideways}\game{Boulderchase}\end{sideways}}
		&\multirow{ 2}{*}{$10^4$}
		&      62.5\%&      75.0\%&\textbf{      82.5\%}&      45.0\%&\textbf{      82.5\%}&      30.0\% \\ 
		&&      23.7&      22.1&      24.0&      18.8&\textbf{      27.3}&      20.1\\\cline{2-8}
		&\multirow{ 2}{*}{$10^3$}
		&      50.0\%&      32.5\%&      37.5\%&\textbf{      52.5\%}&      40.0\%&      22.5\% \\ 
		&&\textbf{      22.8}&      18.6&      18.6&      21.8&      18.1&      16.2\\\cline{2-8}
		&\multirow{ 2}{*}{500}
		&\textbf{      47.5\%}&      30.0\%&      37.5\%&      35.0\%&      32.5\%&      15.0\% \\ 
		&&\textbf{      24.7}&      20.2&      21.4&      18.3&      19.4&      14.4\\\cline{2-8}
		&\multirow{ 2}{*}{250}
		&      40.0\%&      40.0\%&      35.0\%&\textbf{      60.0\%}&      17.5\%&      15.0\% \\ 
		&&      20.9&      20.1&      20.2&\textbf{      21.7}&      14.7&      15.3\\
		\hline
		\hline
		\multirow{8}{*}{\begin{sideways}\game{Whackamole}\end{sideways}}
		&\multirow{ 2}{*}{$10^4$}
		&\textbf{     100\%}&\textbf{     100\%}&\textbf{     100\%}&      75.0\%&      97.5\%&      75.0\% \\ 
		&&\textbf{      72.5}&      44.4&      44.6&      37.0&      60.1&      48.5\\\cline{2-8}
		&\multirow{ 2}{*}{$10^3$}
		&\textbf{     100\%}&\textbf{     100\%}&\textbf{     100\%}&      55.0\%&      77.5\%&      65.0\% \\ 
		&&\textbf{      64.0}&      41.8&      48.2&      33.9&      43.9&      39.0\\\cline{2-8}
		&\multirow{ 2}{*}{500}
		&\textbf{     100\%}&\textbf{     100\%}&\textbf{     100\%}&      57.5\%&      70.0\%&      52.5\% \\ 
		&&\textbf{      59.5}&      50.0&      51.5&      29.0&      38.1&      35.4\\\cline{2-8}
		&\multirow{ 2}{*}{250}
		&      97.5\%&\textbf{     100\%}&      97.5\%&      50.0\%&      65.0\%&      52.5\% \\ 
		&&\textbf{      54.8}&      45.9&      46.4&      28.5&      35.1&      26.6\\
		\hline
		\hline
		\multirow{8}{*}{\begin{sideways}\game{Zelda}\end{sideways}}
		&\multirow{ 2}{*}{$10^4$}
		&\textbf{      97.5\%}&      87.5\%&      90.0\%&      95.0\%&      90.0\%&      70.0\% \\ 
		&&       8.3&       7.4&       6.7&       3.8&\textbf{       9.6}&       8.1\\\cline{2-8}
		&\multirow{ 2}{*}{$10^3$}
		&      80.0\%&      85.0\%&      77.5\%&\textbf{      87.5\%}&      57.5\%&      42.5\% \\ 
		&&\textbf{       8.8}&       7.5&       7.4&       5.3&       8.6&\textbf{       8.8}\\\cline{2-8}
		&\multirow{ 2}{*}{500}
		&      62.5\%&      75.0\%&      70.0\%&\textbf{      77.5\%}&      50.0\%&      35.0\% \\ 
		&&       8.6&       8.2&       7.8&       4.6&\textbf{       8.8}&       7.8\\\cline{2-8}
		&\multirow{ 2}{*}{250}
		&      55.0\%&      55.0\%&      57.5\%&\textbf{      70.0\%}&      45.0\%&      30.0\% \\ 
		&&\textbf{       8.4}&       7.8&       7.8&       4.4&       8.0&       7.2\\
		\hline
		\hline
		\multicolumn{2}{c||}{$\varnothing$ Rank}
		&       1.9
		&       3.1
		&       3.1
		&       3
		&       4.3
		&       5.7
	\end{tabular}
\end{table}

\section{Experimental Results}
\label{sec:Experiments}
\begin{figure}[htb]
	\centering
	\begin{subfigure}{\columnwidth}
		\centering
		\includegraphics[width=0.9\columnwidth]{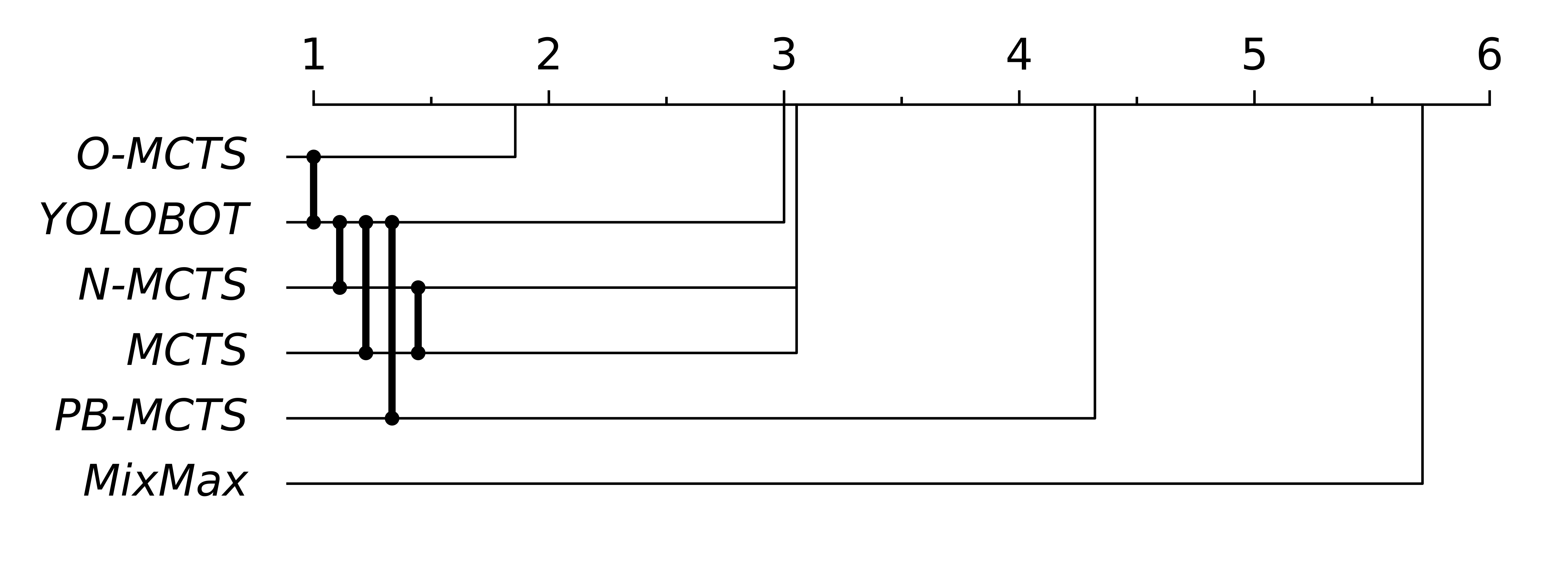}
		\caption{All game runs. Data from Table~\ref{tbl:reslt}}
		\label{fig:wilcoxonAll}
	\end{subfigure}
	
	\begin{subfigure}{\columnwidth}
		\centering
		\includegraphics[width=0.9\columnwidth]{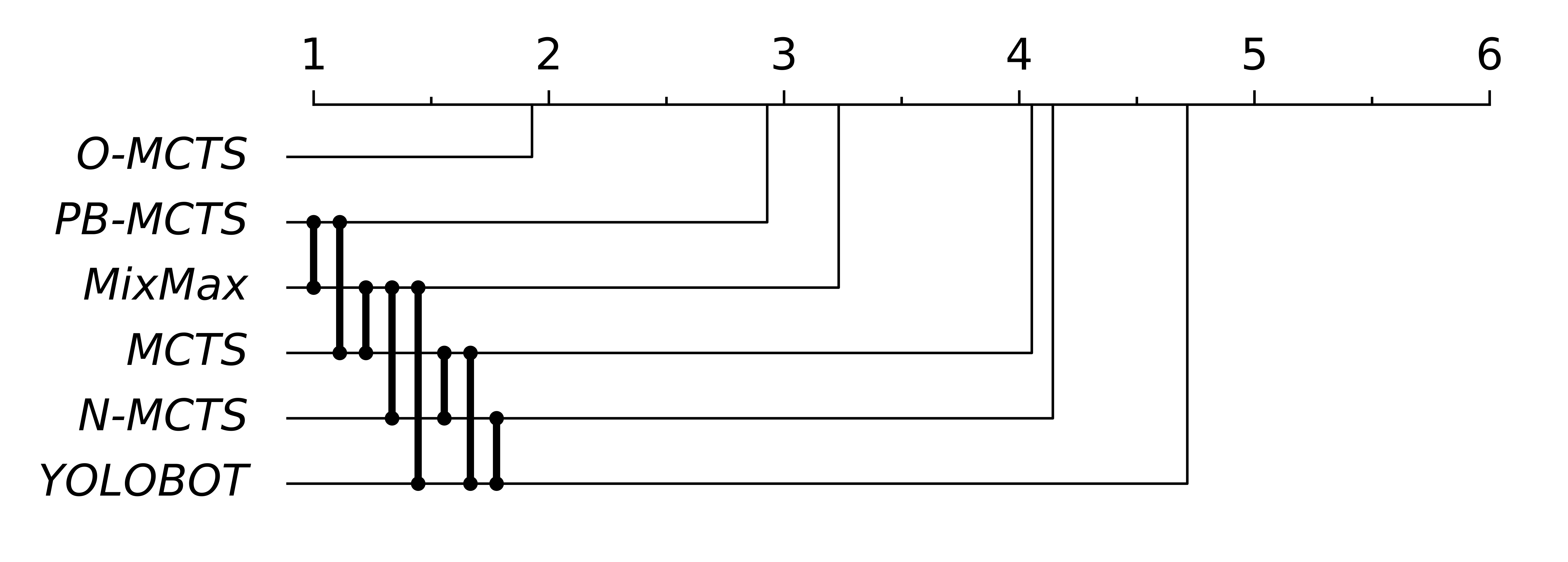}
		\caption{Only won game runs}
		\label{fig:wilcoxonWon}
	\end{subfigure}
	\caption{Average ranks and the result of a Wilcoxon signed rank test with $\alpha=0.01$. Directly connected algorithms do not differ significantly.}
\end{figure}

%
Table~\ref{tbl:reslt} shows the best win rate and the corresponding average score of each algorithm, averaged over $40$ runs
for each of the $36$ different parameter settings.
In each row, the best values for the win rate and the average score are shown in bold, and a ranking of the algorithms is computed. The resulting average ranks 
are shown in the last line. 
We use 
a Friedmann test and a posthoc Wilcoxon signed rank test as an indication for significant differences in performance.
The results of the latter (with a significance level of $99\%$) are shown in Figure~\ref{fig:wilcoxonAll}.

We can see that \omcts{} performed best with an average rank of $1.9$ and a significantly better performance than all other MCTS variants. Only the advanced algorithm \Yolobot{}, which has won the GVGAI competition several times, comes close to it, as 
can be seen in Figure~\ref{fig:wilcoxonAll}.
Table~\ref{tbl:reslt} allows us to take a closer look on the domains where \omcts{} is better:
For games that are easy to win, such as \game{Surround}, \game{Aliens}, and \game{Whackamole} \omcts{} beats the other algorithms MCTS-like algorithms by winning with a higher score.
In \game{Chase}, a deadly but more deterministic game, \omcts{} is able to achieve a higher win rate.
In deadly and stochastic games like \game{Zelda}, \game{Boulderchase} and \game{Jaws} \omcts{} gets beaten by \Yolobot{}, \nmcts{} or MCTS, but still performs good.

\nmcts{} and MCTS perform similarly in all games, which lets us conclude
that per-node normalization does not strongly influence the performance.
\mixmax{} performed worst on nearly every game: In hard games, \mixmax{} does not win often and in high-score games it falls short in score.
In the recorded videos,\footnote{You can watch the videos at \url{https://bit.ly/2ohbYb3}} one can see that \mixmax{} greedily goes for high scores: For example in \game{Zelda}, it approaches enemies where MCTS often flees.
This often leads to a bad rated death.
But nevertheless, \mixmax{} achieves a good score in \game{Zelda} compared to MCTS or \nmcts{}.
In \game{Whackamole}, \mixmax{} dies often most probably because of greedily chosen dangerous moves.

Figure~\ref{fig:wilcoxonWon} summarizes the results when only won games are considered. 
It can be seen, that 
in this case, \mixmax{} is better than MCTS or \nmcts{}, but the difference is not significant.
\omcts{} still performs best, but  
\Yolobot{} falls behind. This is because it is designed to primarily maximize the win rate, not the score. 

In conclusion, we found evidence that \omcts{}'s preference for actions that \emph{maximize win rate} works better than MCTS's tendency to \emph{maximize average performance} for the tested domains.

\paragraph{Parameter Optimization}
In Table~\ref{tbl:reslt2} the overall rank over all parameters for all algorithms are shown.
It is clearly visible that a low rollout length $RL$ improves performance and is more important to tune correctly than the exploration-exploitation trade-off $\Cp$.
Since \Yolobot{} has no parameters, it is not shown (rank $15$).
Except for the extreme case of no exploration ($\Cp=0$), \omcts{} with $RL=5$ is better than any other MCTS algorithm.
The best configuration is \omcts{} with $RL=5$ and $\Cp=1.25$.

\paragraph{Video Demonstrations}
For each algorithm and game, we recorded a video where the agent wins.\footnotemark[2]
In those videos it can be seen that \omcts{} frequently plays actions that lead to a higher score, whereas 
MCTS and \nmcts{} play more safely---often too cautious and averse to risking any potentially deadly effect.


\begin{table}[t!]
\caption{Results for different parameters for all algorithms except of \Yolobot{}. In each cell, the overall rank over all games and time resources is shown. The best configuration per algorithm is highlighted.}
        \centering\includegraphics[width=0.98\columnwidth]{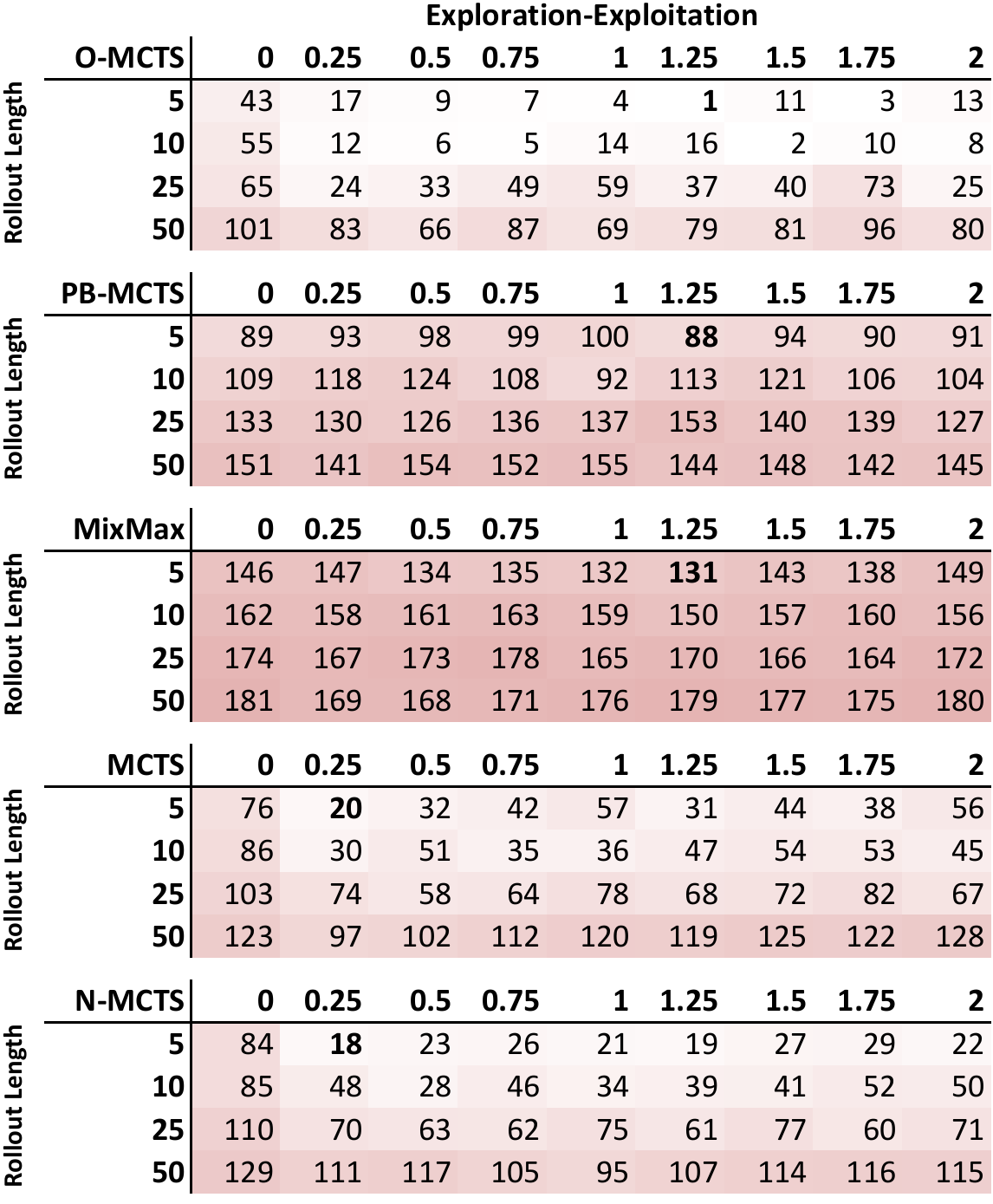}
\label{tbl:reslt2}
\end{table}

\section{Conclusion}
\label{sec:Conclusion}
In this paper we proposed \omcts{}, a modification of MCTS that handles the rewards in an ordinal way:
Instead of averaging backpropagated values to obtain a value estimation, it estimates the winning probability of an action using the Borda score. 
By doing so, the magnitude of distances between different reward signals are disregarded, which can be useful in ordinal domains.
In our experiments 
using
the GVGAI framework, we compared \omcts{} to MCTS, different MCTS modifications and \Yolobot{}, a specialized agent for this domain.
Overall, \omcts{} achieved higher win rates and reached higher scores than the other algorithms, confirming that this approach can be useful in domains where no meaningful numeric reward information is available.

\paragraph{Acknowledgments}
This work was supported by the German Research Foundation (DFG project number FU 580/10). 
We gratefully acknowledge the use of the Lichtenberg high performance computer of the TU Darmstadt for our experiments.

\bibliography{mybib}
\bibliographystyle{aaai}
\end{document}